\documentclass[lettersize,journal]{IEEEtran}
\usepackage{amsmath,amsfonts}
\usepackage{algorithm}
\usepackage{array}
\usepackage[caption=false,font=normalsize,labelfont=sf,textfont=sf]{subfig}
\usepackage{textcomp}
\usepackage{stfloats}
\usepackage{url}
\usepackage{verbatim}
\usepackage{graphicx}
\usepackage{cite}
\newcommand{\eg}{\textit{e.g.}}
\newcommand{\ie}{\textit{i.e.}}
\usepackage{booktabs}
\usepackage{multirow}
\usepackage{multicol}
\usepackage{xcolor}         % colors

\usepackage{algorithmicx}
\usepackage{algpseudocode}
\usepackage[colorlinks,linkcolor=blue]{hyperref}%用于插入超链接
\usepackage[T1]{fontenc}

\begin{document}

\title{Robust Noisy Correspondence Learning via Self-Drop and Dual-Weight}

\author{Fan Liu~\IEEEmembership{Member,~IEEE,} Chenwei Dong, Chuanyi Zhang, Hualiang Zhou, \\Jun Zhou~\IEEEmembership{Senior Member,~IEEE}
        % <-this % stops a space
\thanks{Corresponding author: Chuanyi Zhang.}% <-this % stops a space
\thanks{Fan Liu, Chenwei Dong are with the College of Computer Science and Software Engineer, Hohai University, Nanjing, 210098, China.}
\thanks{Chuanyi Zhang is with the College of Artificial Intelligence and Automation, Hohai University, Nanjing, 210098, China.}
\thanks{Hualiang Zhou is with NARI Technology Co., Ltd., Nanjing, 211106, China}
\thanks{Jun Zhou is with the School of Information and Communication Technology, Griffith University, Nathan, Queensland 4111, Australia.}}

\markboth{Journal of \LaTeX\ Class Files,~Vol.~14, No.~8, August~2021}%
{Shell \MakeLowercase{\textit{et al.}}: A Sample Article Using IEEEtran.cls for IEEE Journals}

\IEEEpubid{0000--0000/00\$00.00~\copyright~2021 IEEE}

\maketitle

\begin{abstract}
Many researchers collect data from the internet through crowd-sourcing or web crawling to alleviate the data-hungry challenge associated with cross-modal matching. Although such practice does not require expensive annotations, it inevitably introduces mismatched pairs and results in a noisy correspondence problem. Current approaches leverage the memorization effect of deep neural networks to distinguish noise and perform re-weighting. However, briefly lowering the weight of noisy pairs cannot eliminate the negative impact of noisy correspondence in the training process. In this paper, we propose a novel self-drop and dual-weight approach, which achieves elaborate data processing by qua-partitioning the data. Specifically, our approach partitions all data into four types: clean and significant, clean yet insignificant, vague, and noisy. We analyze the effect of noisy and clean data pairs and find that for vision-language pre-training models, a small number of clean samples is more valuable than a majority of noisy ones. Based on this observation, we employ self-drop to discard noisy samples to effectively mitigate the impact of noise. In addition, we adopt a dual-weight strategy to ensure that the model focuses more on significant samples while appropriately leveraging vague samples. Compared to the prior works, our approach is more robust and demonstrates relatively more stable performance on noisy datasets, especially under a high noise ratio. Extensive experiments on three widely used datasets, including Flickr30K, MS-COCO, and Conceptual Captions, validate the effectiveness of our approach.

\end{abstract}


\section{Introduction}
\IEEEPARstart{C}{ross-modal} matching~\cite{lee2018stacked,diao2021similarity,liu2020graph, wang2024attribute} endeavors to establish correspondences and alignments between content from distinct modalities, such as images and text or images and audio. Despite the rapid advancements in cross-modal matching methods in recent years, most methods assume that the training data is correctly aligned across various modalities. However, collecting such manual-annotated datasets, \eg, Flickr30K~\cite{young2014image} and MSCOCO~\cite{lin2014microsoft}, is labor-intensive and time-consuming in real-world scenarios. Therefore, to mitigate the high cost of annotation, recent datasets, \eg, Conceptual Captions~\cite{sharma2018conceptual}, were acquired through crowd-sourcing or web crawling. Inevitably, numerous mismatched pairs are erroneously considered as matched ones, leading to the issue of noisy correspondence~\cite{huang2021learning}. In contrast to the extensively researched issue of noisy labels~\cite{han2018co,yu2019does,song2019selfie,yuan2023mcre,ye2020purifynet, wu2018light}, noisy correspondence refers to mismatched cross-modality pairs, posing a more challenging problem than noisy labels as they represent instance-level noise~\cite{jiang2020beyond}.

\begin{figure}[t]
\centering
\includegraphics[width=0.47\textwidth]{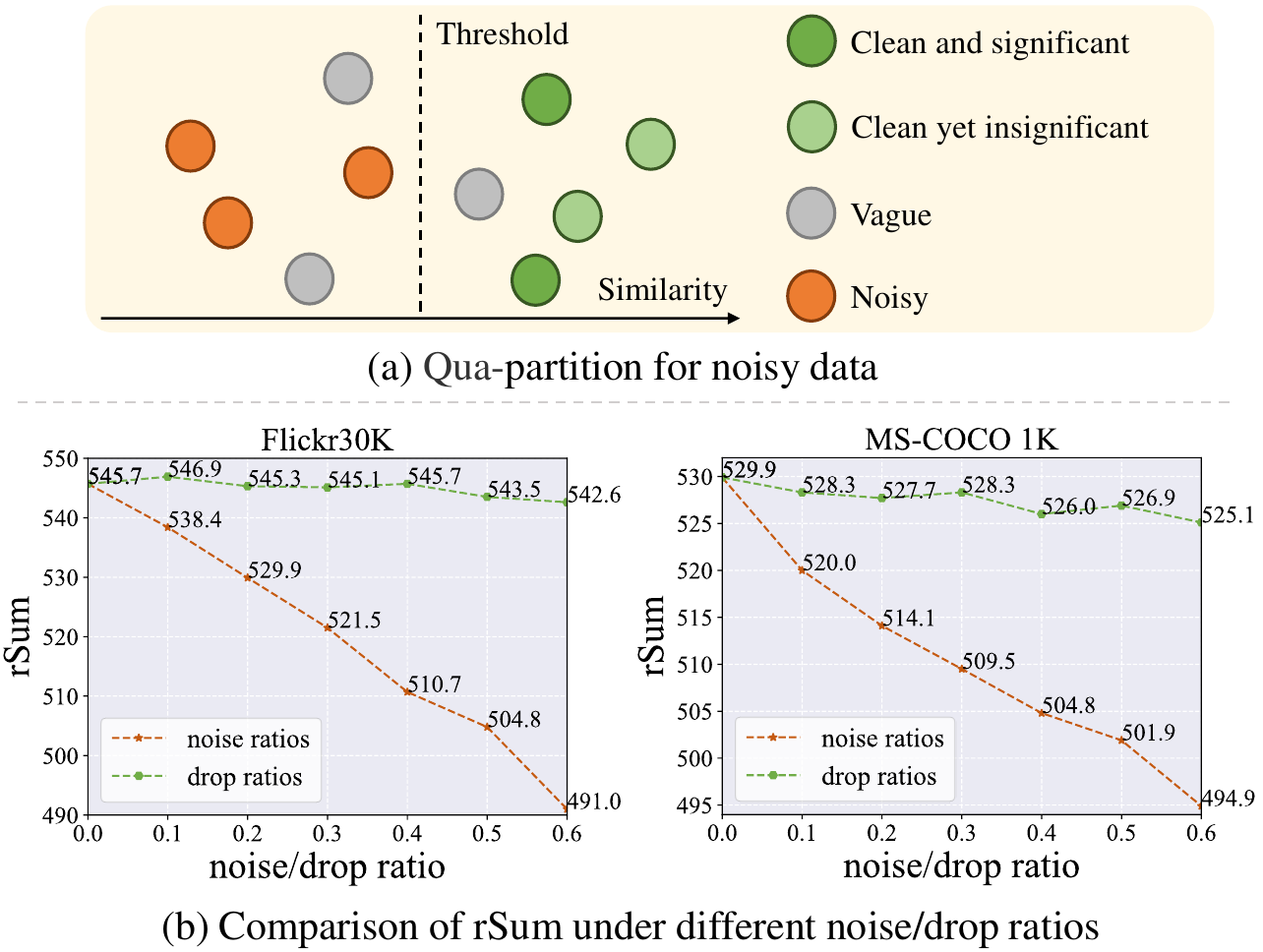} % Reduce the figure size so that it is slightly narrower than the column.
% \captionsetup{font=small}
\caption{Illustrations of complex data distributions in the real world and our observations. (a) We partition all data into four types, namely, clean and significant (green), clean yet insignificant (light green), vague (grey), and noisy (orange). The previous methods of adopting bi-partition or tri-partition are insufficient to handle such complex data distributions. (b) Comparison of rSum of CLIP under different noise/drop ratios on Flickr30K and MS-COCO 1K. The performance of VLP is differently sensitive to noisy data and discarded data and therefore a small number of clean samples is more valuable than a majority of noisy ones for fine-tuning a VLP.}
\label{motivation}
\end{figure}

Many works have paid attention to the issue of noisy correspondence~\cite{huang2021learning,ma2024cross,huang2024learning}.
These methods are typically based on the memorization effect~\cite{arpit2017closer,xia2020robust} of Deep Neural Networks (DNNs) that models tend to learn simple patterns before fitting noisy samples. This effect makes clean pairs display lower losses than noisy ones in the initial few epochs. 
% and training data can be \blue{briefly} separated into two partitions (clean and noisy sets) \blue{through loss values. }
\IEEEpubidadjcol
Therefore, prior works typically train from scratch and employ a warm-up process to achieve preliminary model convergence. Nonetheless, for vision-language pre-training models (VLP) such as CLIP~\cite{radford2021learning}, the cost of training from scratch is substantial. In response to this issue, DSDMR~\cite{shibreaking} employs the remarkable zero-shot capabilities of pre-trained models to differentiate between noisy and clean samples. NPC~\cite{zhang2024negative} evaluates the negative impact of each sample to assign confidence weights. Surprisingly, NPC's performance under noise-free conditions outperforms CLIP, demonstrating that not all clean samples are equally significant for VLP.

Most existing methods typically adopt a bi-partition assumption~\cite{huang2021learning, huang2024learning} that divides data into clean and noisy subsets. Some researchers further investigate tri-partition approaches~\cite{feng2023learning, ma2024cross} and divide data into clean, noisy, and vague subsets, where vague samples are challenging to distinguish from noisy ones.
% \red{Most existing methods, such as bi-partition methods \cite{huang2021learning, huang2024learning}, divide data into clean and noisy subsets, or tri-partition methods \cite{feng2023learning, ma2024cross}, further divide data into clean, noisy, and vague subsets.}
However, real-world data distributions tend to be more complex.
Since clean samples can be unequally significant for VLP, all data can be partitioned into four types: clean and significant, clean yet insignificant, vague, and noisy (Fig.~\ref{motivation}~(a)).
% In practice, as illustrated in Figure~\ref{fig1}(a),
Consequently, simple data partitioning strategies tend to be insufficient for handling complex data distributions and cannot eliminate the impact of noise.

To address the aforementioned challenges, we propose an efficient approach called Self-Drop and Dual-weight (SDD). Our approach is predicated on the observation that a small number of clean samples is more valuable than a majority of noisy ones for fine-tuning VLP. As illustrated in Fig.~\ref{motivation}~(b), we adjust the noise and data drop ratios on Flickr30K and MSCOCO to evaluate the performance of CLIP. CLIP demonstrates relatively stable performance across varying data drop ratios. Conversely, as the noise ratio gradually increases, the model's performance rapidly declines. This contrasting phenomenon becomes more pronounced with increasing noise and drop ratio. It indicates that noisy data is detrimental for VLP while decreasing the amount of training data only has a minor effect. Inspired by this observation, our method aims to identify and discard noisy data, even including some vague samples, to minimize optimization risks. Specifically, our SDD encompasses two core modules to mitigate the impact of noise and enhance its focus on significant parts within clean samples. In the first module, SDD leverages the zero-shot capabilities inherent in VLP to compute the similarity of the given image-text pairs. Subsequently, SDD treats samples with low similarity as noisy ones and opts to discard them. This sample selection strategy can effectively mitigate the impact of noise to benefit robust cross-modal matching.
In the second module, SDD introduces a dual-weight strategy to assign confidence and significance weights from different perspectives. In detail, the confidence weights are dynamically determined by the similarity between sample pairs, and the significance weights are derived by assessing the samples' contribution to model training. The former appropriately leverages vague samples, while the latter ensures the model focuses more on clean and significant samples. 

Our main contributions can be summarized as follows:
\begin{itemize}
    % \item \red{We emphasize that existing works should consider a more complex data assumption, \ie, qua-partition, and highlight the detrimental effects of noisy data and decreasing training data for VLP are not consistent.}
    
    % find that not all samples hold equal significance when fine-tuning VLP.}

    \item We emphasize that existing works should consider a more complex data assumption: clean and significant, clean yet insignificant, vague, and noisy. Properly treating four types of data contributes to noise-robust training.
    
    \item We propose an efficient Self-Drop and Dual-weight (SDD) approach to achieve robustness against noisy correspondence. Specifically, self-drop effectively mitigates the misguidance from noisy data. Dual-weight strategy enhances the impact of clean and significant samples while appropriately leveraging vague ones.
    
    \item Extensive experiments on three widely used datasets, including Flickr30K, MS-COCO, and Conceptual Captions, demonstrate the effectiveness of our approach.
\end{itemize}

\section{RELATED WORK}
In this section, we provide a brief overview of recent advancements in three interconnected fields: cross-modal matching, learning with noisy labels, and learning with noisy correspondence.

\subsection{Cross-modal Matching}
Cross-modal matching strives to create associations and alignments between content from various modalities, such as images and text. Conventional image-text matching models can be broadly classified into two types: 1) global-level matching methods~\cite{faghri2017vse++, liu2017learning, qian2021dual}, which aligned the visual features captured from an image with the overall semantic feature extracted from a text; and 2) local-level matching methods~\cite{lee2018stacked, diao2021similarity, li2019visual, chen2020imram}, which aimed to establish fine-grained connections between local regions of an image and individual words in text.

In recent years, vision-language pre-training models, \eg, CLIP~\cite{radford2021learning}, have demonstrated powerful zero-shot capabilities. Moreover, CLIP has been shown to exhibit flexibility, allowing it to be incorporated into a wide range of cross-modal tasks, including detection~\cite{gu2021open,zhong2022regionclip}, segmentation~\cite{liang2023open}, and captioning~\cite{mokady2021clipcap}. 
% Therefore, we aim to leverage the robust zero-shot capabilities of CLIP to address the issue of noisy correspondence.

However, CLIP still exhibits a high sensitivity to noise when attempting to generalize to downstream tasks through fine-tuning. 
In this paper, we employ CLIP as our backbone and propose a novel approach for fine-tuning VLP for cross-modal matching under the noisy correspondence scenario.

\subsection{Learning with Noisy Labels}
Learning with noisy labels has been extensively studied and is generally concerned with addressing the label noise problem in classification tasks. Noisy label learning algorithms can be broadly categorized into four types: adding regularization~\cite{shorten2019survey, krogh1991simple, srivastava2014dropout, ioffe2015batch}, loss adjustment~\cite{reed2014training, chang2017active}, sample selection~\cite{arazo2019unsupervised,han2018co,yu2019does}, and label correction~\cite{zheng2020error,song2019selfie}. Adding regularization can effectively mitigate overfitting~\cite{song2022learning} such as data augmentation~\cite{shorten2019survey}, weight decay~\cite{krogh1991simple}, dropout~\cite{srivastava2014dropout} and batch normalization \cite{ioffe2015batch}, yet it exhibits suboptimal performance in high-noise scenarios. Loss adjustment is adaptively conducted based on the distinction between noisy and clean samples. Reed et al.~\cite{reed2014training} enhanced the robustness against noise by injecting consistency through network prediction targets. Active Bias~\cite{chang2017active} emphasized samples with uncertain predictions by assigning the predicted variance as the training weight. Sample selection involves choosing clean samples for training from a noisy dataset. Inspired by the memorization effect~\cite{arpit2017closer} of DNNs, Arazo et al.~\cite{arazo2019unsupervised} selected clean samples with the discrepancy in loss values observed during the training process. Although this method is outstanding, it inevitably leads to error accumulation. Therefore, subsequent methods~\cite{han2018co,yu2019does} often employed two DNNs for training in order to mitigate error accumulation. The objective of label correction is to rectify the labels of unreliable samples. AdaCorr~\cite{zheng2020error} introduced a novel label correction algorithm based on the predictions of a noise classifier. SELFIE~\cite{song2019selfie} selectively refurbished and utilized imprecise samples that can be corrected with high accuracy, thereby progressively augmenting the number of available training samples.

However, since learning with noisy labels is tailored for classification tasks and noisy correspondence is an instance-level issue, the methods designed for noisy labels cannot be directly transferred to address the noisy correspondence.

\subsection{Learning with Noisy Correspondence}
Noisy correspondence refers to mismatched pairs that are erroneously considered matched ones. Given the broad applicability of data pairs, methods against noise correspondence have developed across numerous fields, including person re-identification~\cite{yang2022learning, yang2024robust,wu2024modality,qin2024noisy}, graph matching~\cite{lin2023graph}, multi-view clustering~\cite{huang2020partially,yang2021partially,yang2022robust,lu2024decoupled}, and image captioning~\cite{kang2023noise,fu2024noise}.

\begin{figure*}[t]
\centering
\includegraphics[width=1\textwidth]{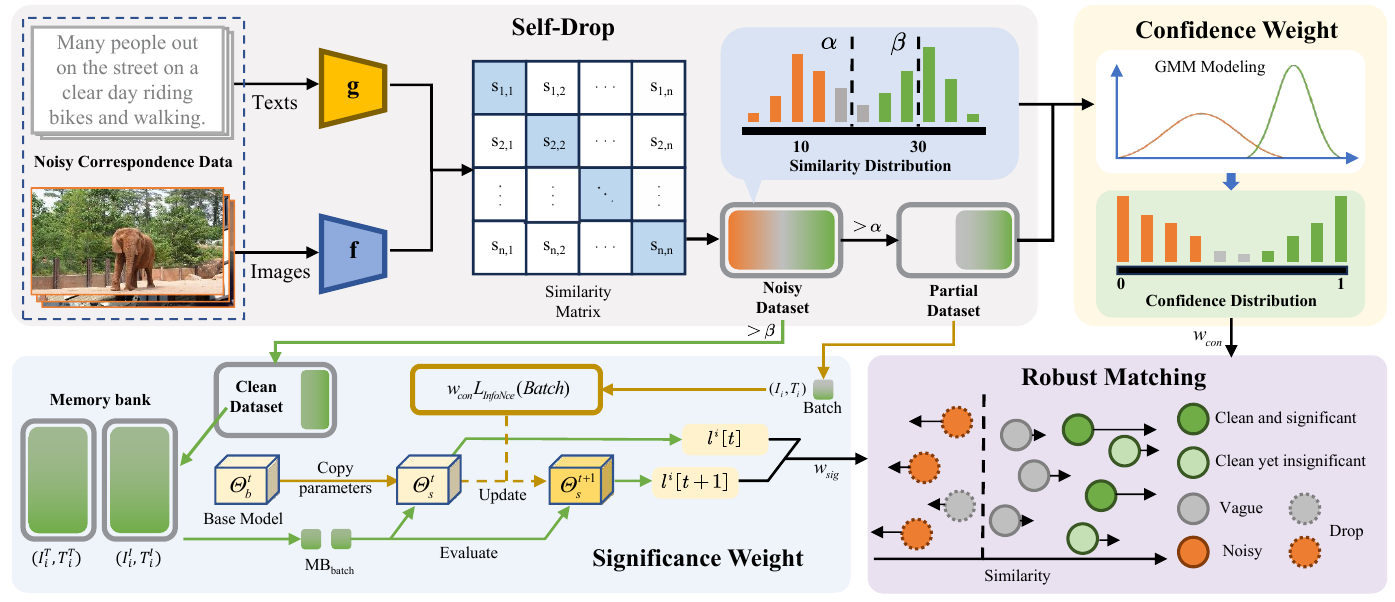} % Reduce the figure size so that it is slightly narrower than the column.
\caption{Overview of the proposed method. \textbf{(1) Self-Drop:} SSD computes the similarity for samples with noisy correspondence, then utilizes a threshold $\alpha$ to construct a partial dataset $D_{p}$ by dropping samples with low similarity. \textbf{(2) Confidence Weight:} GMM generates $w_{con}$ from similarity distribution for samples in the partial dataset. \textbf{(3) Significance Weight:} SDD creates a siamese model by copying the parameters $\Theta_b^t$ of the base model. The siamese model with parameters $\Theta_s^t$ utilizes a memory bank to evaluate the loss variation before ($l^{i}[t]$) and after training ($l^{i}[t+1]$) on the partial dataset to produce $w_{sig}$. \textbf{(4) Robust Matching:} Finally, the base model trains and updates parameters on the partial dataset utilizing $w_{con}$ and $w_{sig}$.}
% \captionsetup{font=small}
\label{overview}
\end{figure*}

Earlier works~\cite{huang2021learning,qin2022deep,hu2023cross} on cross-modal matching have explored the issue of noisy correspondence. NCR~\cite{huang2021learning} recast the rectified labels into soft margins for triplet loss, thereby enabling robust cross-modal matching. CTPR~\cite{feng2023learning} further extended NCR by dividing the image-text pairs within the training data into subsets of clean, hard, and noisy pairs to mitigate the challenge of selecting hard samples. BiCro~\cite{yang2023bicro} reassigned the noisy image-text pairs through bidirectional cross-modal similarity consistency that similar images should have similar textual descriptions and vice versa. MSCN~\cite{han2023noisy} introduced meta-learning and proposed a meta-correction network to provide reliable noise similarity scores. CREAM~\cite{ma2024cross} rectified possible noisy correspondence within positive sample pairs and exploited diverse latent consistency in negative sample pairs. GSC~\cite{zhao2024mitigating} introduced a multimodal geometrical structure consistency by optimizing a contrastive loss that aligns the geometrical structures within modalities and a traditional loss for cross-modal alignment, thereby preserving the integrity of geometrical structures. ESC~\cite{yang2024equivariant} posited that for any two matched samples, the semantic changes induced by image variations should be proportional to those induced by text variations. By leveraging this equivariant similarity consistency, ESC achieves robust cross-modal matching. 

However, the noisy correspondence research in VLP is still in its infancy. DSDMR~\cite{shibreaking} employs the remarkable zero-shot capabilities of pre-trained models to differentiate between noisy and clean samples. NPC~\cite{zhang2024negative} evaluates the negative impact of each sample to assign confidence weights. Although these models have demonstrated promising performances, they typically adopt a bi-partition assumption or tri-partition assumption. In contrast to these works, we adopt a more detailed assumption to achieve robustness against noisy correspondence.

\section{METHOD}
In this section, we provide a detailed explanation of the proposed SDD, which consists of two modules integrated with a novel objective function to achieve robust cross-modal matching against noisy correspondences. Section~\ref{sec:sec3A} defines the problem of noisy correspondence. Section~\ref{sec:sec3B} revisits the feature representation in CLIP. Section~\ref{sec:sec3C} discusses the method of discarding the majority of noisy samples through self-drop. Section~\ref{sec:sec3D} elaborates on the dual-weight strategy, which adaptively weights the remaining samples based on confidence and importance. Section~\ref{sec:sec3E} introduces the proposed objective function to robust cross-modal matching. Finally, we discuss the differences between our SDD and NPC~\cite{huang2021learning} in Section~\ref{sec:sec3F} to show our innovation. The framework of SDD is illustrated in Fig.~\ref{overview}.

\subsection{Problem Definition}\label{sec:sec3A}
To ensure broad applicability, we explore the issue of noisy correspondence in cross-modal matching with the example of image-text matching. Given the noisy dataset $D = \{(I_i, T_i)\}_{i=1}^N$ that consists of a total of $N$ training samples, where $(\,I_i, T_i)\,$ represents the $i$-th image-text pair. Generally, the aim of image-text matching models with parameters $\Theta$ is to project visual and textual modalities into a shared representation space via image encoder $f$ and text encoder $g$, respectively. Then the similarity of the given image-text pairs is calculated through $S(I_i,T_i)$, which we abbreviate as $s_{i,i}$ for simplicity, as expressed in the following equation:
\begin{equation}\label{eq1}
    S(I_i,T_i)=s_{i,i}=\frac{f(I_i) \cdot g(T_i)}{||f(I_i)|| \cdot ||g(T_i)||}.
\end{equation}

\begin{algorithm*}[t]
\caption{Self-Drop and Dual-Weight Training Algorithm}
\label{alg1}
\textbf{Input}: Training set $D=\{(I_i, T_i)\}_{i=1}^N$, memory bank $D_{MB}=\{(I_i^I, T_i^I), (I_i^T, T_i^T)\}_{i=1}^N$\\
% \textbf{Parameter}: Optional list of parameters\\
\textbf{Output}: Robust matching model $\theta_b^E$
\begin{algorithmic}[1] %[1] enables line numbers
\State Initialize parameters $\theta_b^e$ for base model
\For{each epoch $e=1,2,...,E$}
    \State $//$ \textit{self-drop}
    \State Compute the similarity of the given image-text pairs $s=\{s_{i,i}\}_{i=0}^N$ by Eq. (\ref{eq1})

    \State Construct partial dataset $D_p=\{(I_i, T_i)|s_{i,i}>\alpha\}$
    \State $//$ \textit{confidence weight}
    \State Obtain confidence weights $\{w_{con}^i\}_{i=0}^N\longleftarrow GMM(s)$ by Eq. (\ref{eq5})-(\ref{eq6})
    \State Copy parameters from $\theta_b^e$ to create siamese
model $\theta_s^e$
    \For{each mini-batch $\{(I_i, T_i)\}_{i=1}^B$ from $D_p$ with a batch size of $B$}
        \State $//$ \textit{significance weight}
        \State Compute siamese model loss $l_{i}=\{(l_{i2t}^i[e],l_{t2i}^i[e])\}_{i=1}^B$ by Eq. (\ref{eq9})
        \State Calculate siamese model loss $w_{con}^i\mathcal L_\mathrm{InfoNCE}(I_i,T_i)$ with mini-batch $\{(I_i, T_i)\}_{i=1}^B$
        \State Update siamese model parameters $\theta_s^e$ to $\theta_s^{e+1}$
        \State Compute siamese model loss $l_{i}=\{(l_{i2t}^i[e+1],l_{t2i}^i[e+1])\}_{i=1}^B$ by Eq. (\ref{eq9})
        \State Calculate significance weights $\{w_{sig}^i\}_{i=1}^B$ by Eq. (\ref{eq10})-(\ref{eq11})
        \State $//$ \textit{robust matching}
        \State Calculate base model loss $w_{con}^iw_{sig}^i\mathcal L_\mathrm{InfoNCE}(I_i,T_i)$ with mini-batch $\{(I_i, T_i)\}_{i=1}^B$
        \State Update base model parameters $\theta_b^e$ to $\theta_b^{e+1}$
    \EndFor
\EndFor
\end{algorithmic}
\end{algorithm*}

\subsection{Revisiting Feature Representations in CLIP}\label{sec:sec3B}
In various vision-language learning tasks, the VLP model CLIP~\cite{radford2021learning} has demonstrated remarkable generalization capabilities.
% VLP model CLIP \cite{radford2021learning} has demonstrated impressive generalization ability in various vision-language learning tasks. 
In this paper, we choose CLIP as the backbone of our approach. It adopts a simple InfoNCE loss function for optimization~\cite{oord2018representation}, which promotes the alignment of matched image-text samples while pushing apart mismatched samples:
\begin{equation}\label{eq2}
    \mathcal L_\mathrm{InfoNCE}(I_i,T_i)=\mathcal CE(I_i,T_i)+\mathcal CE(T_i,I_i),
\end{equation}

\begin{equation}\label{eq3}
    \mathcal CE(x_i, y_i)=-\log\left( \frac{\exp (S(x_i, y_i)/ \tau)}{\begin{matrix}\sum_{j=1}^{B} \end{matrix} \exp(S(x_i, y_j)/ \tau)}\right),
\end{equation}
where $\tau$ is a temperature parameter and $B$ is the batch size.

CLIP achieves strong zero-shot capabilities by leveraging large-scale image-text pairs collected from the internet for training. However, as shown in Fig.~\ref{motivation}~(b), CLIP still exhibits a high sensitivity to noise when attempting to generalize to downstream tasks through fine-tuning.

\subsection{Self-Drop Module}\label{sec:sec3C}
% In this section, we elaborate on the proposed SDD which consists of two modules: Self-Drop and Dual-weight.
Prior works are typically based on the memorization effect~\cite{arpit2017closer,xia2020robust} of DNNs, which tend to learn simple patterns before fitting noisy samples. Consequently, these works typically train from scratch and employ a warm-up process to achieve preliminary model convergence. However, this paradigm presents two shortcomings: 1) for VLP, such as CLIP, the cost of training from scratch is substantial; and 2) in the warm-up period, models would inevitably fit the noisy samples~\cite{huang2024learning}, thus degrading the performance. To tackle these issues, by considering the strong zero-shot capabilities inherent in CLIP, we aim to identify and discard noisy data, even including vague samples, to minimize optimization risk.

To begin, as illustrated in Fig.~\ref{overview}, we project $N$ image-text pairs into a shared embedding space through image and text encoder, respectively. Subsequently, we compute the cosine similarity of image-text pairs by Eq.~\eqref{eq1} and store it in a matrix of size $N^2$. The required similarity of image-text pairs is located on the diagonal of the matrix, which can be expressed as $\{S_{i,i}\}_{i=1}^N$. Since noisy samples are detrimental (Fig.~\ref{motivation}~(b)), selecting an appropriate proportion of noisy data and discarding them may be a more cost-effective approach compared to utilizing the noisy data. Specifically, by setting a threshold $\alpha$ for $\{s_{i,i}\}_{i=1}^N$, we drop the majority of noisy pairs to construct the partial dataset. 

% \red{It is important to note that the cosine function clamps similarity within [-1,1], resulting in smaller differences between positive and negative pairs \cite{jiang2024negative}. For example, the cosine similarity for positive pairs in CLIP is typically around 0.3, while for negative pairs it is around 0.1 \cite{ming2022impact}. Due to the minimal difference, CLIP typically employs a temperature parameter to amplify it by a factor of 100. Thus, we can infer that in CLIP, the similarity between noisy pairs is approximately 10, while the similarity between clean pairs is around 30. For simplicity, we set the threshold $\alpha$ to 20 through all experiments. The samples with $s_{i,i} \le \alpha$ will be discarded and remaining samples will construct a partial dataset as follows:}
As a special phenomenon of the CLIP-based model, the cosine similarity of positive pairs in CLIP is typically around 0.3, while of negative pairs it is around 0.1~\cite{jiang2024negative,ming2022impact}. Due to the minimal difference, CLIP usually employs a temperature parameter $\tau$ to amplify it by a factor of 100. Thus, In CLIP, the similarity scores for noisy and clean pairs are approximately 10 and 30, respectively. For simplicity, we set the threshold $\alpha$ to 20 through all experiments. The selection of hyperparameter $\alpha$ will be discussed in subsequent visualization experiments. Samples with $s_{i,i} \le \alpha$ will be discarded and the remaining samples will construct a partial dataset as follows:
\begin{equation}\label{eq4}
    D_p=\{(I_i, T_i)|s_{i,i}>\alpha\}.
\end{equation}

\subsection{Dual-Weight Module}\label{sec:sec3D}
After self-drop, it is still risky to directly train the model on $D_p$~\cite{dang2024noisy} because noise discarding cannot be perfect.
% after discarding the majority of noisy pairs,
The filtered dataset $D_p$ potentially contains a small number of noisy pairs and vague pairs that are difficult to distinguish from the noisy ones~\cite{feng2023learning, ma2024cross}. Additionally, clean yet insignificant samples are often overlooked by prior works. Therefore, we propose a dual-weight strategy that assigns confidence and significance weights from different perspectives to suppress noise data, properly utilize vague pairs, and enhance the contribution of significant clean samples.
\subsubsection{Confidence Weight} The purpose of confidence weight is to minimize the impact of a small number of noisy samples in $D_p$ while appropriately leveraging the vague samples within $D_p$. Following DSDMR~\cite{shibreaking}, we fit the similarity of all pairs by a two-component Gaussian Mixture Model (GMM)~\cite{permuter2006study, li2020dividemix}:
\begin{equation}\label{eq5}
    p(x|\theta)=\sum_{k=1}^{K}\alpha_k\phi(x|\theta_k),
\end{equation}
where $\alpha_k$ denotes the mixture coefficient, and $\phi(x|\theta_k)$ denotes the probability density of the $k^{th}$ component. The confidence probability $w_{con}^i$ of pair $i$ is calculated by:
\begin{equation}\label{eq6}
    w_{con}^i=p(\theta_k|x_i)=\frac{p(\theta_k)p(x_i|\theta_k)}{p(x_i)},
\end{equation}
where $\theta_k$ is the Gaussian component with the higher mean.

We utilize $w_{con}$ generated by the GMM to re-weight the data in $D_p$. After that, vague samples can contribute to model training with proper weight, while the negative impact of a small portion of noisy samples in $D_p$ is diminished.

\subsubsection{Significance Weight}
NPC's~\cite{zhang2024negative} performance on noise-free remarkably outperforms CLIP, demonstrating that not all clean samples are equally significant for VLP. Inspired by this observation, we introduce a significant weight into our approach.

It is a rational assumption that the model's performance should be improved after training on clean samples compared to before the training. In other words, if the model's performance is worse after training than before, we can regard training samples that are insignificant or even noisy.
% conclude that these samples are clean but insignificant.
For convenience, we utilize the loss value of the model as a substitute for its performance.
We evaluate the change in loss values before and after training with a memory bank $(MB)$ to calculate the significance of each sample. The memory bank is a reliable set of clean samples that correspond to each sample in the noisy dataset $D$.

To obtain $MB$, a clean dataset $ D_{c} $ is sampled with a relatively high threshold $\beta$ (30):
% \red{Samples in a partial dataset $D_p$ are evaluated to obtain significance weight by absolutely clean samples from the memory bank.}
% Unlike self-drop, we opt for a higher threshold $\beta$ (30) to ensure the acquisition of a strict clean set:
\begin{equation}\label{eq7}
    D_c=\{(I_i, T_i)|s_{i,i}>\beta\}.
\end{equation}
Then, for each image-text pair $(I_i,T_i)$ in the dataset $D$, we assign two evaluation entries $(I_i^I, T_i^I)$ and $(I_i^T, T_i^T)$ to it, where $I_i^I$ shows the highest similarity with $I_i$ in $D_c$ and $T_i^T$ shows the highest similarity with $T_i$ in $D_c$. Specifically, the memory bank can be denoted as:
\begin{equation}\label{eq8}
    D_{MB}=\{(I_i^I, T_i^I), (I_i^T, T_i^T)\}_{i=1}^N, (I_i^*,T_i^*) \in D_{c}.
\end{equation}

Samples in the partial dataset $D_p$ are evaluated to obtain significance weight by clean samples from the memory bank.
As illustrated in Fig.~\ref{overview}, we initially create a siamese model by copying the parameters $\Theta_b$ of the base model, with the parameters of the siamese model represented as $\Theta_s$. We use the model's loss $ l^{i} $ on clean $D_{MB} $ to represent performance, where a lower loss indicates a higher significance of the sample. Formally, for the $i$-th data pair $(I_i, T_i)$ in $D_p$, we can compute its \textit{i2t} loss and \textit{t2i} loss of its evaluation entry in the current epoch $e$ as follows:
\begin{equation}\label{eq9}
    \begin{split}
        &l_{i2t}^i[e]=\mathcal CE(I_i^I, T_i^I)+\mathcal CE(I_i^T,T_i^T),\\
        &l_{t2i}^i[e]=\mathcal CE(T_i^I, I_i^I)+\mathcal CE(T_i^T, I_i^T).
    \end{split}
\end{equation}
Then, we train the siamese model using data from $D_p$ weighted by $w_{con}$ and update the siamese model parameters $\Theta_s^e$ to $\Theta_s^{e+1}$. The change in performance after updating the siamese model parameters can be calculated by:
\begin{equation}\label{eq10}
    c_i=\frac{1}{2}(\frac{l_{i2t}^i[e]}{l_{i2t}^i[e+1]}+\frac{l_{t2i}^i[e]}{l_{t2i}^i[e+1]}).
\end{equation}
Finally, the significance of clean samples can be defined by:
\begin{equation}\label{eq11}
    w_{sig}^i=
    \begin{cases}
    tanh(c_i)&, \text{if $c_i<1$}\\
    1&,\text{otherwise}
    \end{cases}.
\end{equation}
The values of performance change $c_i$ are densely distributed around 1. Therefore, we aim to discretize the weights of significant and insignificant samples through a mapping function, such as the $tanh$ or the $sigmoid$ function.
% Specifically, in our research, we employ the $tanh$ as our chosen mapping.
Samples that decrease performance after training ($c_i<1$) will be assigned with smaller weights, while those that improve performance ($c_i\geq1$) are set the weights to 1.

\subsection{Loss Function}\label{sec:sec3E}
% Finally, we train model on partial dataset $D_{p}$.
% With the significance weight $w_{sig}$ produced by the siamese model with $\theta_s^{t+1}$, we can update the base model parameters $\theta_b^t$ to $\theta_b^{t+1}$ by utilizing Dual-weight.
After obtaining dual-weight $w_{con}$ and $w_{sig}$, we can update the base model parameters $\Theta_b^e$ to $\Theta_b^{e+1}$ by utilizing partial dataset $D_{p}$.
The overall loss function of SDD is formulated as expressed as follows:
\begin{equation}\label{eq12}
    \mathcal L=\sum_{(I_i,T_i) \in D_{p}} w_{con}^{i}w_{sig}^{i}\mathcal L_\mathrm{InfoNCE}(I_i,T_i),
\end{equation}
where confidence weight $w_{con}^{i}$ minimizes the impact of noisy samples and appropriately leverages vague samples, while significance weight $w_{sig}^{i}$ ensures the model focuses more on clean and significant samples. Algorithm~\ref{alg1} summarizes our proposed SDD. 

\subsection{Discussions}\label{sec:sec3F}
It should be acknowledged that NPC~\cite{zhang2024negative} is one of our motivations as well as a SOTA approach for noisy correspondence problems. In this subsection, we present detailed discussions from various aspects to deeply recognize the differences between our SDD and NPC. We believe a comprehensive analysis could provide a more thorough understanding of our approach.

\subsubsection{Sample Selection}
NPC aims to estimate the negative impact of each sample and re-weights all samples before model training without sample selection. Conversely, our proposed SDD is established from the observation that a small number of clean samples is more valuable than a majority of noisy ones for fine-tuning VLP. Therefore, we perform sample selection to explicitly discard a large proportion of noisy data to enhance the data reliability.

% \begin{algorithm}[t]
% \caption{Construct Memory Bank}
% \label{alg2}
% \textbf{Input}: Training set $D=\{(I_i, T_i)\}_{i=1}^N$\\
% % \textbf{Parameter}: Optional list of parameters\\
% \textbf{Output}: memory bank $D_{MB}=\{(I_i^I, T_i^I), (I_i^T, T_i^T)\}_{i=1}^N$
% \begin{algorithmic}[1] %[1] enables line numbers
% \State Initialize parameters $\theta_b^t$ for model

% \State Compute sample similarity: $\{s_{i,i}\}_{i=0}^N$
% \State Construct clean dataset $D_c=\{(I_i, T_i)|s_{i,i}>\beta\}$
% \For{each sample $(I_i, T_i)$ from $D$ }
%     \State Search for the most similar $I_i^I$ to $I_i$ from clean dataset $D_c$ and construct evaluation entry $(I_i^I, T_i^I)$
%     \State Search for the most similar $T_i^T$ to $T_i$ from clean dataset $D_c$ and construct evaluation entry $(I_i^T, T_i^T)$
% \EndFor

% \end{algorithmic}
% \end{algorithm}

\subsubsection{Weighting Method}
Our significance weight is indeed motivated by NPC, indicating that not each sample contributes equally to model training. However, in our research, we discover that relying solely on a single weight (NPC's method) is insufficient to eliminate the negative impact of noise (we will demonstrate it in experiment Fig.~\ref{r1_compare}). Therefore, we further explore the weighting strategy and propose a dual-weight method to re-weight samples from both confidence and significance perspectives.

\begin{table*}[!ht]
 	\caption { Image-text matching performance under synthetic noise ratios of 20\%, 40\%, and 60\% on Flickr30K and MS-COCO 1K. The \textbf{best} and \underline{second-best} results are highlighted in each column.}
  \centering
	\newcommand{\tabincell}[2]{\begin{tabular}{@{}#1@{}}#2\end{tabular}}
		\begin{tabular}{c|c|ccc|ccc|c|ccc|ccc|c}
			%\toprule
		\toprule
		&&\multicolumn{7}{c|}{Flickr30K}&\multicolumn{7}{c}{MS-COCO 1K}\\
		&&\multicolumn{3}{c|}{Image$\longrightarrow$Text}&\multicolumn{3}{c|}{Text$\longrightarrow$Image}&&\multicolumn{3}{c|}{Image$\longrightarrow$Text}&\multicolumn{3}{c|}{Text$\longrightarrow$Image}&\\
		% \midrule
  Noise&Methods&R@1&R@5&R@10&R@1&R@5&R@10&rSum&R@1&R@5&R@10&R@1&R@5&R@10&rSum\\
			\midrule
			\multirow{9}{*}{20\%}
                % &SCAN& 56.4   & 81.7  & 89.3 & 34.2 & 65.1  & 75.6 & 402.3  & 28.9 & 64.5 & 79.5 & 20.6 & 55.6 & 73.5 & 322.6\\

			% ~&SAF & 51.8   & 79.5 & 88.3 & 38.1 & 66.8 & 76.6  & 401.1 & 41.0 & 78.4  & 89.4 & 38.2 & 74.0 & 85.5 & 406.5 \\

			~&NCR~\cite{huang2021learning} & 75.0   & 93.9 & 97.5 & 58.3 & 83.0 & 89.0 & 496.7 & 77.7 & 95.6 & 98.2 & 62.6 & 89.3 & 95.3  & 518.7  \\

			~&BiCro~\cite{yang2023bicro} &78.1&94.4&97.5&60.4&84.4&89.9&504.7&78.8&96.1&98.6&63.7&90.3&95.7&523.2\\

            ~&MSCN~\cite{han2023noisy}&76.4&94.5&97.6&58.8&83.5&89.2&500.0&78.1&\textbf{97.2}&98.8&64.3&90.4&95.8&524.6\\

            ~&CRCL~\cite{qin2023cross}&77.9&95.4&98.3&60.9&84.7&90.6&507.8&79.6&96.1&98.7&64.7&90.6&95.9&525.6\\

            ~&GSC~\cite{zhao2024mitigating}&78.3&94.6&97.8&60.1&84.5&90.5&505.8&79.5&96.4&\underline{98.9}&64.4&90.6&95.9&525.7\\

            ~&ESC~\cite{yang2024equivariant}&79.0&94.8&97.5&59.1&83.8&89.1&503.3&79.2&\underline{97.0}&\textbf{99.1}&64.8&90.7&96.0&526.8\\

            ~&DSDMR~\cite{shibreaking}&85.1&\underline{97.0}&\underline{99.2}&69.7&90.3&94.8&536.1&\underline{80.5}&95.8&98.4&65.8&\underline{90.9}&96.2&527.6\\

            ~&NPC~\cite{zhang2024negative}&\underline{87.3}&\textbf{97.5}&98.8&\underline{72.9}&\underline{92.1}&\underline{95.8}&\underline{544.4}&79.9&95.9&98.4&\underline{66.3}&90.8&\underline{98.4}&\underline{529.7}\\
            ~&\textbf{SDD}&\textbf{87.6}&\textbf{97.5}&\textbf{99.5}&\textbf{74.3}&\textbf{93.4}&\textbf{96.8}&\textbf{549.1}&\textbf{81.4} & 96.0 & 98.5 & \textbf{67.1} & \textbf{91.2 }& \textbf{98.5} & \textbf{532.7}\\
			\midrule
			\multirow{9}{*}{40\%}
                % &SCAN& 29.9    & 60.5 & 72.5 & 16.4 & 38.5 & 48.6 & 266.4 & 30.1 & 65.2 & 79.2 & 18.9 & 51.1 & 69.9 & 314.4 \\

			% ~&SAF & 34.3   & 65.6 & 78.4 & 30.1 & 58.0 & 68.5  & 335.0 & 36.0 & 74.4 & 87.0 & 33.7  & 69.4  & 82.5 & 383.0 \\
			~&NCR~\cite{huang2021learning} & 73.5   & 92.6 & 95.8 & 55.7 & 80.3 & 86.9 & 484.8 & 76.6 & 95.6 & 98.2  & 61.0 & 88.9 & 94.9 & 515.2 \\
			~&BiCro~\cite{yang2023bicro}&74.6&92.7&96.2&55.5&81.1&87.4&487.5&77.0&95.9&98.3&61.8&89.2&94.9&517.1\\
			
           ~&MSCN~\cite{han2023noisy}&69.5&90.8&95.7&53.2&79.9&86.4&475.5&74.5&\underline{96.0}&98.1&60.8&89.0&95.0&513.4\\
           
            ~&CRCL~\cite{qin2023cross}&77.8&95.2&98.0&60.0&84.0&90.2&505.2&78.2&95.7&98.3&63.3&90.3&95.7&521.5\\

            ~&GSC~\cite{zhao2024mitigating}&76.5&94.1&97.6&57.5&82.7&88.9&497.3&78.2&95.9&98.2&62.5&89.7&95.4&519.9\\
           
            ~&ESC~\cite{yang2024equivariant}&76.1&93.1&96.4&56.0&80.8&87.2&489.6&78.6&\textbf{96.6}&\textbf{99.0}&63.2&\underline{90.6}&95.9&523.9\\
           
           ~&DSDMR~\cite{shibreaking}&85.2&97.0&\underline{99.0}&69.6&90.2&94.9&535.9&\underline{79.9}&95.9&98.3&64.8&90.3&95.9&525.1\\
           ~&NPC~\cite{zhang2024negative}&\underline{85.6}&\underline{97.5}&98.4&\underline{71.3}&\underline{91.3}&\underline{95.3}&\underline{539.4}&79.4&95.1&98.3&\underline{65.0}&90.1&\underline{98.3}&\underline{526.2}\\
           ~&\textbf{SDD}&\textbf{87.3}&\textbf{98.0}&\textbf{99.2}&\textbf{73.5}&\textbf{92.9}&\textbf{96.6}&\textbf{547.5}& \textbf{81.5} &95.9 &\underline{98.5} &\textbf{66.8} &\textbf{91.0} &\textbf{98.5} &\textbf{532.2} \\
			\midrule
			\multirow{9}{*}{60\%}
                % &SCAN& 16.9     & 39.3 & 53.9 & 2.8  & 7.4 & 11.4 & 131.7 & 27.8 & 59.8 & 74.8 & 16.8  & 47.8  & 66.4  & 293.4   \\

			% ~&SAF  & 28.3      & 54.5  & 67.5  & 22.1  & 47.3  & 59.0  & 278.7   & 28.2  & 63.9  & 79.4  & 31.1  & 65.6  & 80.5  & 348.7  \\

			~&NCR~\cite{huang2021learning}  & 70.0     & 91.0 & 94.4 & 52.3   & 76.9 & 84.0 & 468.6 & 72.6  & 93.8  & 97.4  & 57.0  & 86.4  & 93.6    & 500.8   \\
			
            ~&BiCro~\cite{yang2023bicro}&67.6&90.8&94.4&51.2&77.6&84.7&466.3&73.9&94.4&97.8&58.3&87.2&93.9&505.5\\

            ~&MSCN~\cite{han2023noisy}&68.8&88.6&93.1&48.8&76.4&84.0&459.7&73.7&95.1&\underline{98.5}&57.0&86.9&94.0&505.2\\

            ~&CRCL~\cite{qin2023cross}&73.1&93.4&95.8&54.8&81.9&88.3&487.3&76.3&95.1&97.9&60.8&89.0&95.1&514.2\\      
			
            ~&GSC~\cite{zhao2024mitigating}&70.8&91.1&95.9&53.6&79.8&86.8&478.0&75.6&95.1&98.0&60.0&88.3&94.6&511.7\\   
            
            ~&ESC~\cite{yang2024equivariant}&72.6&90.9&94.6&53.0&78.6&85.3&475.0&77.2&95.1&98.1&61.1&88.6&94.9&515.0\\
            
            ~&DSDMR~\cite{shibreaking}&\underline{85.8}&\textbf{96.9}&\textbf{99.1}&\underline{69.5}&\underline{90.1}&\underline{94.7}&\underline{536.1}&\underline{78.9}&\underline{95.2}&98.2&\underline{63.5}&\underline{89.5}&95.7&\underline{521.0}\\
            ~&NPC~\cite{zhang2024negative}&83.0&95.9&98.6&68.1&89.6&94.2&529.4&78.2&94.4&97.7&63.1&89.0&\underline{97.7}&520.1\\
            ~&\textbf{SDD}&\textbf{87.2}&\underline{96.8}&\underline{99.0}&\textbf{72.4}&\textbf{92.4}&\textbf{96.2}&\textbf{544.0}&\textbf{79.5}&\textbf{95.9}&\textbf{98.6}&\textbf{65.2}&\textbf{90.3}&\textbf{98.6}&\textbf{528.1}\\
			
			%\hline
			% Baseline (CE)              & 33.8  & 52.8  & 35.4 & \textbf{19.6} & 38.7  & 44.2 %\\
			
			\bottomrule
	\end{tabular}
        % \captionsetup{font=small}

 \label{table1}
\end{table*}

\section{Experiments}
In this section, we present a comprehensive experimental validation of the proposed SDD from multiple perspectives. To this end, we conducted extensive experiments on image-text retrieval across three benchmark datasets. The structure of this section is organized as follows. In Section~\ref{sec:sec4A}, we provide a detailed description of the experimental setup, including the datasets and implementation details. In Section~\ref{sec:sec4B}, we compare the performance of our method with 8 state-of-the-art (SOTA) approaches to validate its effectiveness. Furthermore, we assess the noise robustness of our method through stability comparisons and progressive analyses. Finally, in the interest of fairness, we also compare SDD with other methods that similarly utilize a CLIP ViT-B/32-based backbone. In Section~\ref{sec:sec4C}, we conduct a series of ablation studies to provide a comprehensive understanding of SDD.

\subsection{Experimental Setting}\label{sec:sec4A}
\subsubsection{Datasets and Evaluation Metrics}
Following previous works, the proposed SDD was evaluated on three benchmark datasets, Flickr30K~\cite{young2014image}, MSCOCO~\cite{lin2014microsoft}, and CC120K~\cite{zhang2024negative}:

\begin{itemize}
    \item Flickr30K comprises 31,783 images with 5 annotated texts per image. Following previous works~\cite{huang2021learning}, 29,783, 1,000, and 1,000 images were used for training, validation, and testing, respectively.
    \item MS-COCO encompasses 123,287 images with 5 annotated captions per image. Following previous works~\cite{huang2021learning}, we employed 113,287 images for training, with 5,000 images allocated for validation and 5,000 images reserved for testing.
    \item CC120K is a subset of the web-crawled dataset Conceptual Captions~\cite{sharma2018conceptual}, with about 3\%-20\% mismatched image-text pairs. CC120K consists of 120,851 images with a single caption for each. Following NPC~\cite{zhang2024negative}, we utilized 118,851 images for training, 1,000 images for validation, and 1,000 images reserved for testing.

    % \red{For fair comparison, we follow the experimental setup by NPC \cite{zhang2024negative} and select a subset from the realworld dataset Conceptual Captions \cite{sharma2018conceptual}.} This dataset is collected from the Internet, with about 3\%-20\% mismatched image-text pairs. CC120K consists of 120,851 images with a single caption for each. In our experiment, we \red{utilize} 118,851 images for training, \red{1,000 images for validation and with an additional 1,000 images reserved for testing.}

\end{itemize}

\begin{table}[ht]
\caption{Image-Text Matching on MS-COCO 5K. The \textbf{Best} and \underline{second-best} results are highlighted.}
  \centering
 \setlength{\tabcolsep} {0.45mm}
\begin{tabular}{c|c|ccc|ccc|c}
\toprule
   & & \multicolumn{3}{c|}{Image$\longrightarrow$Text} & \multicolumn{3}{c|}{Text$\longrightarrow$Image} & \\
\rule[-1ex]{0pt}{3.5ex} Noise & Methods & R@1 & R@5 & R@10 & R@1 & R@5 & R@10 & rSum \\
\midrule
  \multirow{7}{*}{40\%}
  & NCR~\cite{huang2021learning} &55.5 & 82.4 & 90.2 & 39.7 & 68.5 & 79.2 & 415.5 \\
  & BiCro~\cite{yang2023bicro}& 56.3 &83.0 & 90.8 & 40.1 & 69.0 & 79.5 & 418.7 \\
  & MSCN~\cite{han2023noisy} & 49.7 & 78.9 & 88.0 & 36.9 & 66.1 & 77.1 & 396.7 \\
  & CRCL~\cite{qin2023cross} & 55.8 & 83.1 & 90.1 & 40.9 & 67.8 & 80.6 & 418.3 \\
  & ESC~\cite{yang2024equivariant} & 56.2 & 83.2 & \underline{90.7} & 41.0 & 69.5 & 79.8 & 420.4 \\
  & NPC~\cite{zhang2024negative} & \underline{61.1} & \underline{84.8} & \underline{90.7} & \underline{44.7} & \underline{72.1} & \underline{81.7} & \underline{435.1} \\
  & \textbf{SDD} & \textbf{63.7} & \textbf{85.6} & \textbf{91.8} & \textbf{47.0} & \textbf{73.8} & \textbf{83.0} & \textbf{444.9} \\
  \midrule
    \multirow{7}{*}{60\%}
  & NCR~\cite{huang2021learning} &49.6 & 78.1 & 87.3 & 35.5 & 64.2 & 75.7 & 390.4 \\
  & BiCro~\cite{yang2023bicro}& 52.5 &80.0 & 88.4 & 37.8 & 66.2 & 77.1 & 402.0 \\
  & MSCN~\cite{han2023noisy} & 48.1 & 76.0 & 85.5 & 34.5 & 63.5 & 75.1 & 382.7 \\
  & CRCL~\cite{qin2023cross} & 53.1 & 81.2 & 89.0 & 37.6 & 66.3 & 77.4 & 404.6 \\
  & ESC~\cite{yang2024equivariant} & 53.4 & 81.1 & 89.2 & 38.2 & 66.7 & 77.5 & 406.1 \\
  & NPC~\cite{zhang2024negative} & \underline{59.7} & \underline{82.9} & \underline{89.7} & \underline{43.0} & \underline{70.2} & \underline{79.9} & \underline{425.4} \\
    &\textbf{SDD} & \textbf{62.0} & \textbf{85.1} & \textbf{91.5} & \textbf{45.5} & \textbf{72.2} & \textbf{81.9} & \textbf{438.2} \\
\bottomrule
\end{tabular}
% \captionsetup{font=small}

\label{table2}
\end{table}

\begin{table}[t!]
 \setlength{\tabcolsep} {1mm}
  	\caption{Comparison with baselines on CC120K. The \textbf{Best} and \underline{second-best} results are highlighted in each column.}
  \centering
		\begin{tabular}{c|ccc|ccc|c}
			\toprule
			& \multicolumn{3}{c|}{Image$\longrightarrow$Text}&\multicolumn{3}{c|}{Text$\longrightarrow$Image}&\\
			% \hline
			Methods&R@1&R@5&R@10&R@1&R@5&R@10&rSum\\
			\hline
			CLIP~\cite{radford2021learning}&68.8&87.0&92.9&\underline{67.8}&86.4&90.9&493.8\\
			NPC~\cite{zhang2024negative}&\underline{71.1}&\underline{92.0}&\textbf{96.2}&\textbf{73.0}&\underline{90.5}&\underline{94.8}&\underline{517.6}\\
			\textbf{SDD}&\textbf{72.2}&\textbf{92.2}&\underline{95.5}&\textbf{73.0}&\textbf{91.4}&\textbf{95.0}&\textbf{519.3}\\
			
			\bottomrule
		\end{tabular}
	% \captionsetup{font=small}

  \label{table3}
  \end{table}

We evaluated SDD with the recall rate at K (R@K) which measures the proportion of relevant items found within the top K results of a ranked list. We took the image and text as queries, respectively, and calculated the corresponding results of R@1, R@5, and R@10, which were further summed to evaluate the overall performance, \ie, rSum. Additionally, we used the variance $(var)$ of rSum under different noise ratios to evaluate the approaches’ performance stability, with lower $var$ indicating higher stability.

\subsubsection{Implementation Details} 
We adopted the pre-trained CLIP~\cite{radford2021learning} with ViT-B/32 as our backbone and trained SDD on a single RTX 3090 GPU. We employed a batch size of 128 and an AdamW~\cite{loshchilov2017decoupled} optimizer with a weight decay of 0.2. In all experiments, we trained the model for 5 epochs with a learning rate of $2e-7$. The hyperparameters $\alpha$ and $\beta$ were set to 20 and 30, respectively.

\subsection{Comparison with State-of-the-Art}\label{sec:sec4B}

\subsubsection{Experiments on Flickr30K and MS-COCO}
To verify the effectiveness of SDD, we conducted comparison experiments with 8 state-of-the-art methods, including NCR~\cite{huang2021learning}, BiCro~\cite{yang2023bicro}, MSCN~\cite{han2023noisy}, CRCL~\cite{qin2023cross}, GSC~\cite{zhao2024mitigating}, ESC~\cite{yang2024equivariant}, DSDMR~\cite{shibreaking}, and NPC~\cite{zhang2024negative}. Given that the data in Flickr30K and MS-COCO are correctly paired, we introduced noisy correspondences by randomly shuffling the captions of 20\%, 40\%, and 60\% of training images. Table~\ref{table1} shows results on Flickr30K and MS-COCO under different noise ratios. For Flickr30K and MS-COCO 1K, we obtained results from ESC~\cite{yang2024equivariant} and the publications introducing the respective models. For MS-COCO 5K, we used the data reported in the ESC as the results.

From the experimental results, it can be seen that the proposed SDD demonstrates significant improvements over all state-of-the-art methods. In the case of Flick30K, SDD shows advantages over the second-best method in the rSum score of recalls by 4.7\%, 8.1\%, and 7.9\% under different noise ratios, respectively. For MS-COCO 1K, the rSum scores of SDD are 3.0\%, 6.0\%, and 7.1\% higher than the second-best method. In Table~\ref{table2}, we also demonstrate the superior performance of SDD on the MS-COCO full 5K dataset. SDD surpasses the second-best method in the rSum score of recalls by 9.8\% and 12.8\%.

\subsubsection{Experiments on CC120K} Table~\ref{table3} shows the results on CC120K under a real-world noisy correspondence scenario. The baseline results are derived from NPC~\cite{zhang2024negative} for convenience. It can be observed from Table~\ref{table3} that SDD achieves the best performance with an overall score of 519.3\%, surpassing the second-best method NPC by 1.7\%. The results affirm SDD’s capability to manage not only simply simulated but also complex, real-world noisy correspondences.

% \begin{figure*}[ht!]
% \centering
% \includegraphics[width=1\textwidth]{figure/NPC_distribution.pdf} % Reduce the figure size so that it is slightly narrower than the column.
% \caption{We visualize the similarity distribution of clean and noisy pairs at different training stages of NPC, which is conducted on Flickr30K under 40\% noise.}
% \label{NPC_distribution}
% \end{figure*}

% \begin{figure}[t]
% \centering
% \includegraphics[width=0.47\textwidth]{figure/proportion_noisy.pdf} % Reduce the figure size so that it is slightly narrower than the column.
% % \captionsetup{font=small}
% \caption{The proportion of noisy samples to clean samples in $D_c$ across different noise ratios on Flickr30K.}
% \label{proportion_noisy}
% \end{figure}

% \begin{figure}[t]
% \centering
% \includegraphics[width=0.47\textwidth]{figure/discard_clean.pdf} % Reduce the figure size so that it is slightly narrower than the column.
% % \captionsetup{font=small}
% \caption{The proportion of clean samples among the discarded samples on Flickr30K and MS-COCO 1K.}
% \label{discard_clean}
% \end{figure}

\subsubsection{Stability Comparison}
\begin{figure}[t]
\centering
\includegraphics[width=0.47\textwidth]{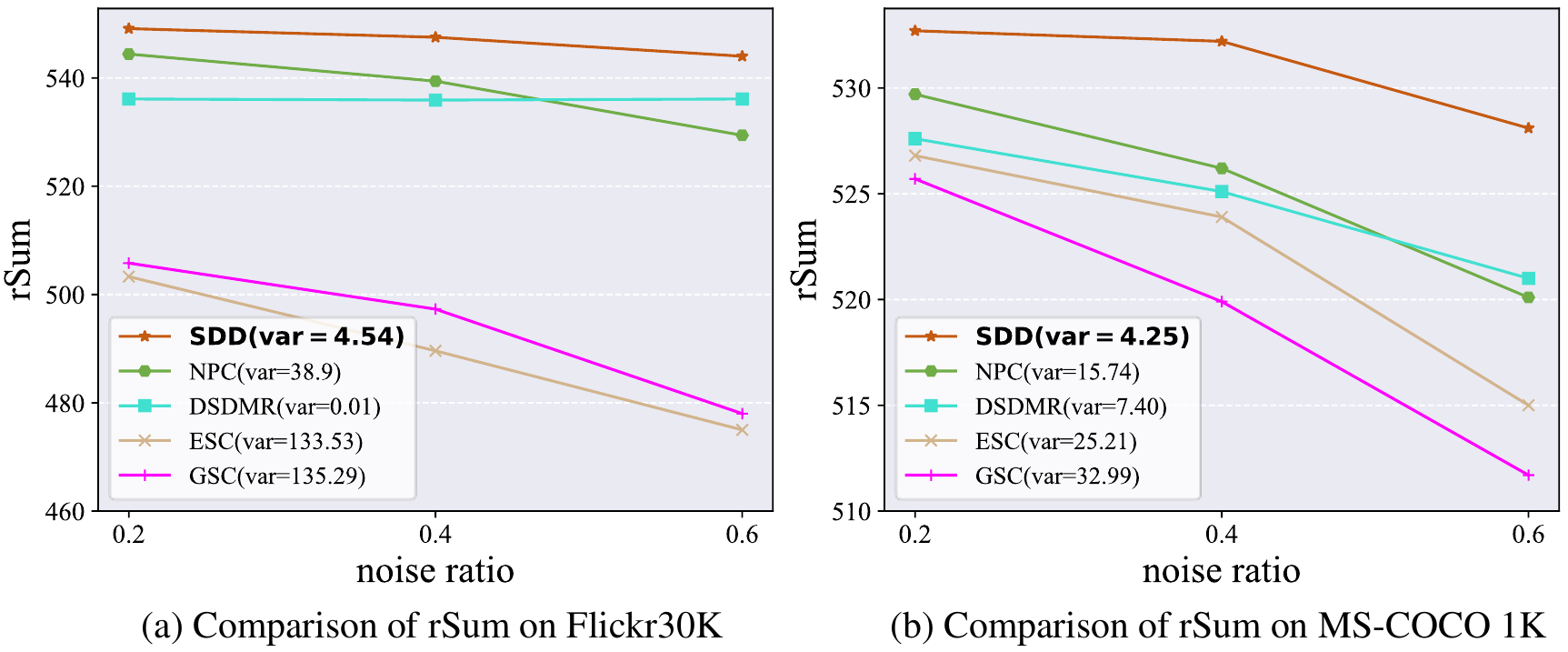} % Reduce the figure size so that it is slightly narrower than the column.
% \captionsetup{font=small}
\caption{(a) Performance and variance $(var)$ of under different noise ratios on Flickr30K. (b) Performance and variance $(var)$ of under different noise ratios on MS-COCO 1K.}
% \caption{(a) Performance and variance $(var)$ of under different noise ratios. (b) Performance curves on the validation set of Flickr30K in the training process.}
\label{rsum_compare}
\end{figure}

To further investigate the performance stability of SDD, we present in Fig.~\ref{rsum_compare} the rSum change curves of different methods on Flickr30K and MS-COCO 1K under varying noise ratios.
% \red{In this experiment, we obtained results from the publications introducing the respective models.} 
The performance of each method is obtained from its original paper respectively.
From Fig.~\ref{rsum_compare}, it can be observed that the proposed SDD considerably outperforms all methods across all noise ratios. SDD demonstrates relatively stable performance under different noise ratios, whereas the performance of other methods significantly decreases as the noise ratio increases. Furthermore, we calculate the variance of each method under different noise ratios to quantify the stability of the methods. 
% Compared with all other methods, our model demonstrates a remarkable stability gap over other approaches with a only 4.25\% variance. 
Our model shows a significant stability gap compared with most other methods, with variances of only 4.54\% and 4.25\% on Flickr30K and MS-COCO 1K, respectively.
Although DSDMR has a lower variance than SDD on Flickr30K, its performance is evidently worse than ours.
These results validate the effectiveness of mitigating the negative effects of noise through self-drop and confidence weight $w_{con}$.

\begin{table}[t]
\centering
\caption{Comparison of methods with ViT-B/32 backbone on noisy MSCOCO. The \textbf{best} and \underline{second-best} results are highlighted in each column.}
\label{table5}%
% \resizebox{0.95\columnwidth}{24mm}{%
\begin{tabular}{c|c|ccc}
\hline
noise                 & method       & 1K R@1        & 5K R@1         & 1K RSUM        \\ \hline
\multirow{7}{*}{20\%} & CLIP~\cite{radford2021learning}         & 66.8          & 47.2           & 507.2           \\
            & VSE$\infty$~\cite{chen2021learning}          & 72.0            & 51.4           & 520.2          \\
                      & PCME~\cite{chun2021probabilistic}      & 69.9          & 48.1           & 519.3          \\
                      & PCME++~\cite{chun2024pcmepp}       & 70.8          & 49.5           & 522.4          \\
                      & PAU~\cite{li2024prototype}       & 71.4             & 51.7           & 521.5              \\
                      & NPC~\cite{zhang2024negative} & \underline{73.1} & \underline{53.8}  & \underline{529.8} \\ 
                      & \textbf{SDD} & \textbf{74.3}      & \textbf{55.5}      & \textbf{532.7} \\ \hline
\multirow{7}{*}{50\%} 
                      & CLIP~\cite{radford2021learning}         & 60.9          & 41.4             & 486            \\
                      & VSE$\infty$~\cite{chen2021learning}          & 38.5          & 18.4           & 390.5          \\
                      & PCME~\cite{chun2021probabilistic}      & 65.8            & 43.0           & 505.7          \\
                      & PCME++~\cite{chun2024pcmepp}       & 65.7          & 44.0             & 503.9          \\
                      & PAU~\cite{li2024prototype}       & 69.3             & 49.6           & 513.4              \\
                      & NPC~\cite{zhang2024negative} & \underline{71.3} & \underline{51.9} & \underline{523.4} \\
                      & \textbf{SDD} & \textbf{73.0} & \textbf{54.5} & \textbf{530.1} \\  \hline
\end{tabular}%
% }
\end{table}

\begin{figure}[t]
\centering
\includegraphics[width=0.47\textwidth]{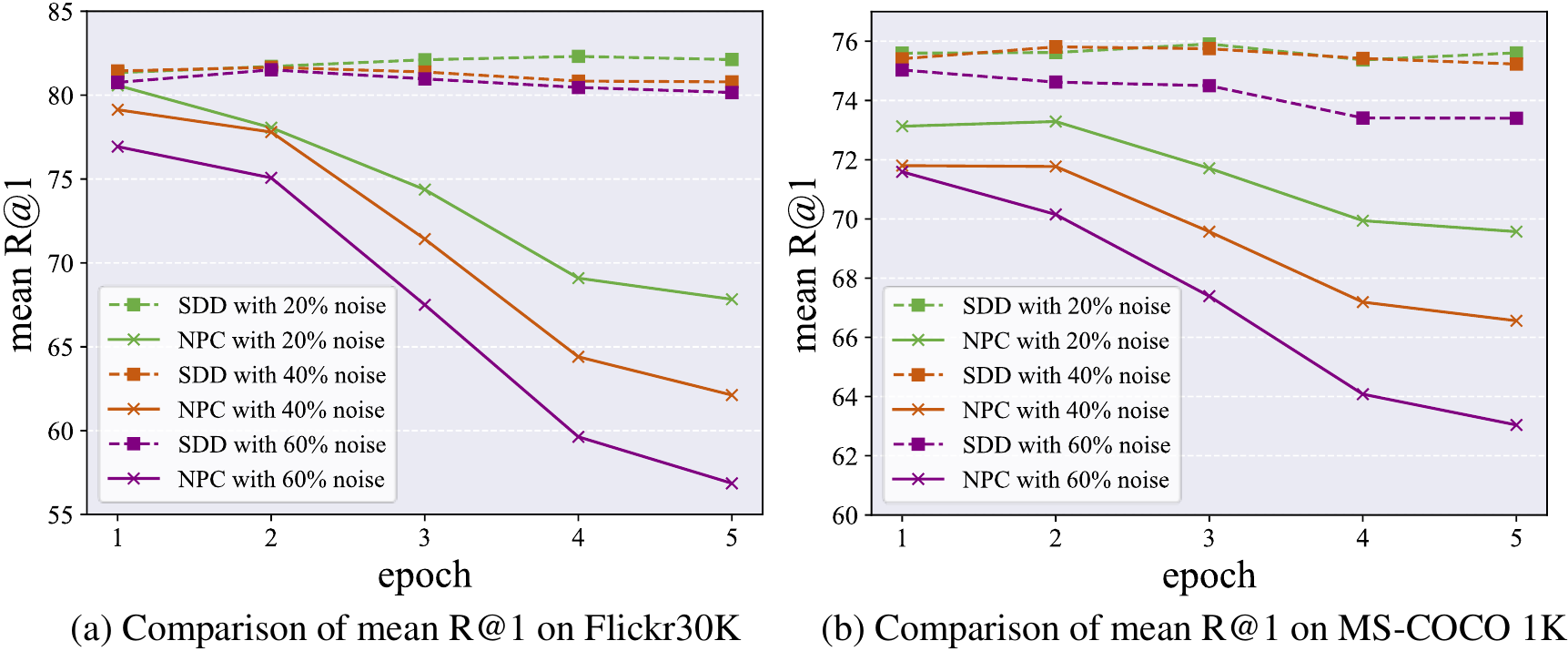} % Reduce the figure size so that it is slightly narrower than the column.
% \captionsetup{font=small}
\caption{(a) Performance curves on the validation set of Flickr30K in the training process. (b) Performance curves on the validation set of MS-COCO 1K in the training process.}
% \caption{(a) Performance and variance $(var)$ of under different noise ratios. (b) Performance curves on the validation set of Flickr30K in the training process.}
\label{r1_compare}
\end{figure}

\begin{figure*}[t]
\centering
\includegraphics[width=1\textwidth]{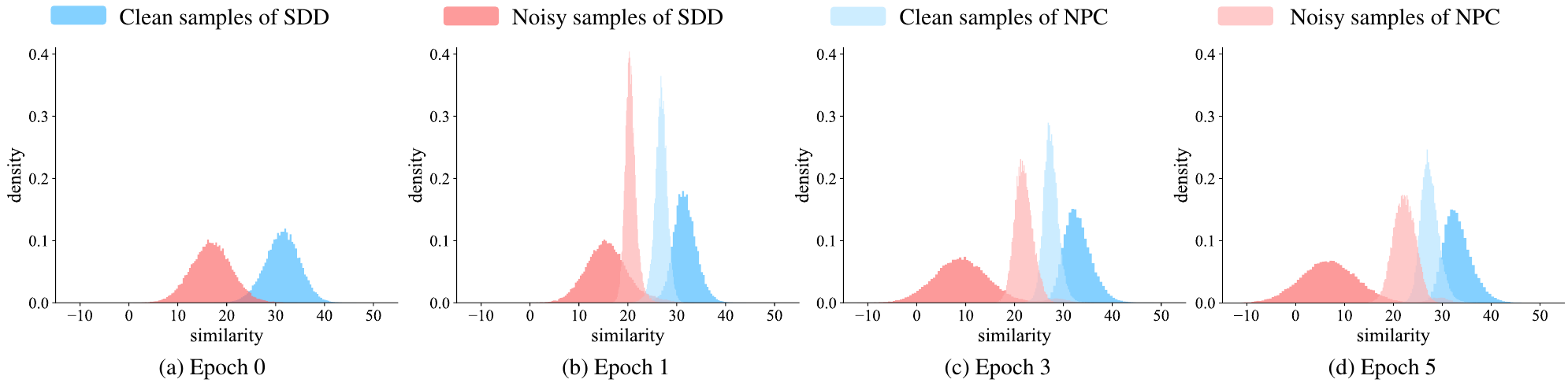} % Reduce the figure size so that it is slightly narrower than the column.
% \captionsetup{font=small}
\caption{Illustration of similarity distributions of SDD and NPC at different training stages on the Flickr30K under 40\% noise. Since SDD and NPC adopt the same CLIP ViT-B/32-based backbone, their similarity distributions are the same in the initial epoch~(a).
% The first column represents the initial stage, the second column shows after training for 1 epoch, the third column shows after training for 3 epochs, and the last column shows after training for 5 epochs. 
Although SDD inevitably fits the noise after training for 5 epochs, it still performs well in separating clean and noisy samples. On the contrary, NPC's separation of noise after 5 epochs~(d) of training is even worse than that in epoch 1~(b).}
\label{visualization_distribution}
\end{figure*}

\begin{table}[t!]
  \centering
    \caption{Effectiveness of each component on Flickr30K with 40\% noise ratio. The best results are marked by \textbf{bold}.}
     % \renewcommand\arraystretch{.75}
    % \small
    \setlength{\tabcolsep} {1.7mm}
    \begin{tabular}{ccc|ccc|c}
    \toprule
          \multicolumn{3}{c|}{Components}      & \multicolumn{4}{c}{Image$\longrightarrow$Text} \\
    \midrule
    self-drop  &  $w_{con}$ & $w_{sig}$ & R@1   & R@5   & R@10 &rSum \\
    \midrule
    $\checkmark$ &   $\checkmark$    &   $\checkmark$    & \textbf{\ 87.3 } & \textbf{\ 98.0 } & 99.2 & \textbf{284.5} \\
     $\checkmark$ & $\checkmark$ &       & 87.2  & 97.8  & 99.1 & 284.1\\
    $\checkmark$ &       & $\checkmark$ & 86.2  & 97.9  & \textbf{99.4}& 283.5 \\
     & $\checkmark$ & $\checkmark$ & 86.7  & \textbf{98.0}  & 99.3& 284.0 \\
    $\checkmark$ &  &  & 85.8  & 97.8  & \textbf{99.4}& 283.0 \\
    \midrule
          &       &       & \multicolumn{4}{c}{Text$\longrightarrow$Image} \\
    $\checkmark$ & $\checkmark$ & $\checkmark$ &  73.5  & \textbf{\ 92.9 } & \textbf{\ 96.6 } & \textbf{263.0}\\
    $\checkmark$ & $\checkmark$ &       & \textbf{73.6}  & 92.8  & 96.3&262.7\\
    $\checkmark$ &       & $\checkmark$ & 72.1  & 92.4  & 96.2 &260.7 \\
     & $\checkmark$ & $\checkmark$ & 73.2  & \textbf{92.9}  & 96.4& 262.5 \\
    $\checkmark$ &  &  & 72.3  & 92.5  & 96.2 & 261.0 \\
    \bottomrule
    \end{tabular}%
    % \captionsetup{font=small}

  \label{table4}%
\end{table}%

\begin{figure}[t]
\centering
\includegraphics[width=0.47\textwidth]{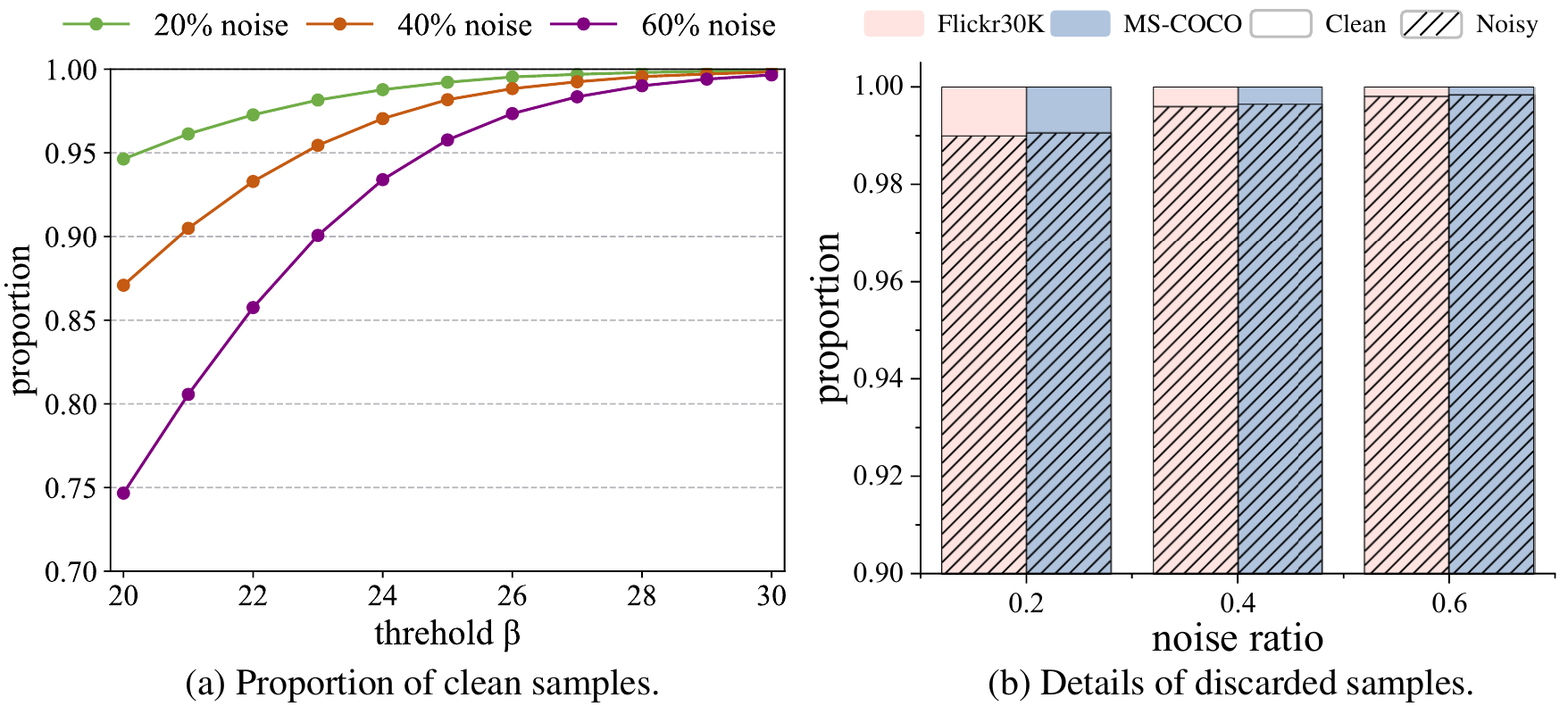}
\caption{(a) The proportion of clean samples in clean dataset $D_c$ across different noise ratios on Flickr30K. (b) The proportion of clean and noisy samples among the discarded samples on Flickr30K and MS-COCO 1K.}
\label{proportion_noisy_and_discard_clean}
\end{figure}

\subsubsection{Progressive Comparison} In order to demonstrate the noise-robustness of our approach, we further investigate the training process by conducting a progressive comparison. We selected NPC as a compared baseline method because it generally demonstrated the second-best performance on three datasets. The results of the NPC were reproduced by us with the same experimental setting as ours.
In Fig.~\ref{r1_compare}, we present the results of the R@1 average values for NPC and SDD on the Flickr30K and MS-COCO validation sets. The results indicate that our SDD demonstrates promising performance across all noise ratios and maintains relative stability throughout the training process. Conversely, although NPC initially exhibits notable performance, its effectiveness gradually declines due to the accumulation of errors.
This phenomenon indicates that using a single weight for re-weighting (NPC's strategy) is insufficient to address the challenges posed by noisy correspondence.
Therefore, we opt to discard noisy samples using self-drop and further mitigate the impact of noise by dual-weight.

\subsubsection{Comparison with ViT-B/32 Backbone Methods}
Since the CLIP ViT-B/32-based backbone inherently has advantages over other backbones, we also provide results of other methods adopting the ViT-B/32 backbone in Table~\ref{table5}, including VSE$\infty$~\cite{chen2021learning}, PCME~\cite{chun2021probabilistic}, PCME++~\cite{chun2024pcmepp}, PAU~\cite{li2024prototype}, NPC~\cite{zhang2024negative}, for a fair comparison. The results of the models compared with SDD in this experiment are all from the paper of NPC~\cite{zhang2024negative}. Specifically, we conducted comparative experiments under 20\% and 50\% noise correspondence scenarios, and report the mean R@1 MS-COCO 1K and 5K, as well as rSum on 1K.

As demonstrated in Table~\ref{table5}, under 20\% and 50\% noise ratios, the rSum of SDD outperforms the previous best-performing method NPC by 2.9\%,  and 6.7\%, respectively. Moreover, under 20\% noise ratio, SDD surpasses NPC by 1.2\% and 1.7\% on MSCOCO 1K and 5K, respectively, and under 50\% noise ratio, SDD surpasses NPC by 1.7\% and 2.6\% on MSCOCO 1K and 5K, respectively. It is worth noting that, as a representative of VLP, CLIP exhibits strong zero-shot capabilities but its performance significantly deteriorates in noisy scenarios. Specifically, when the noise ratio increases from 20\% to 50\%, CLIP's rSum on MS-COCO 1K decreases by 21.2\%. Compared to other CLIP-based methods, our work significantly reduces the risk of learning noise data with CLIP ViT-B/32-based backbone as much as possible, which demonstrates the value of our work.

\subsection{Ablation Study}\label{sec:sec4C}
We undertook the ablation study on the Flickr30K with a noise ratio of 40\% to detailedly demonstrate our approach.

\subsubsection{Effectiveness of each component}
We show the effect of each component in Table~\ref{table4}. Specifically, we ablated the contributions of three key components of SDD, i.e., self-drop, confidence weight $w_{con}$, and significance weight $w_{sig}$. From Table~\ref{table4}, we observe the following conclusions: 1) The full SDD could achieve the best overall performance, showing that all three components are important to improve the robustness against noisy correspondence. 2) The model's performance significantly declines without confidence weight $w_{con}$ demonstrating that a small number of noisy samples can have a severe impact on the model.

\begin{figure*}[t]
\centering
\includegraphics[width=1\textwidth]{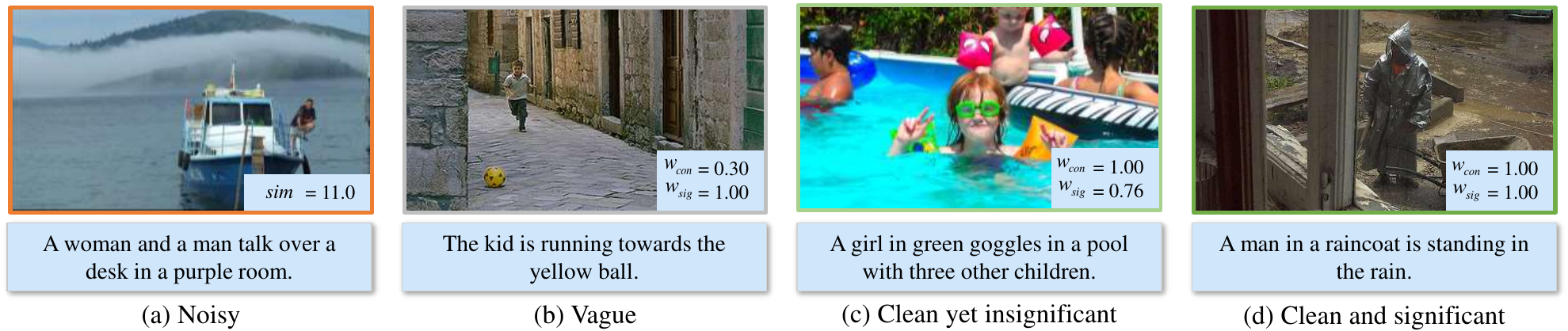} % Reduce the figure size so that it is slightly narrower than the column.
% \captionsetup{font=small}
\caption{Case study on Flickr30K training set under 40\% noise. 
% The noisy sample is marked in orange (a), the vague sample in gray (b), the clean yet insignificant sample in light green (c), and the clean and significant sample in green (d).
The last three samples are assigned with confidence weight $w_{con}$ and significance weight $w_{sig}$, whereas the noisy sample is only labeled with similarity as it is directly discarded.}
\label{case_study}
\end{figure*}

\begin{figure*}[t]
\centering
\includegraphics[width=1\textwidth]{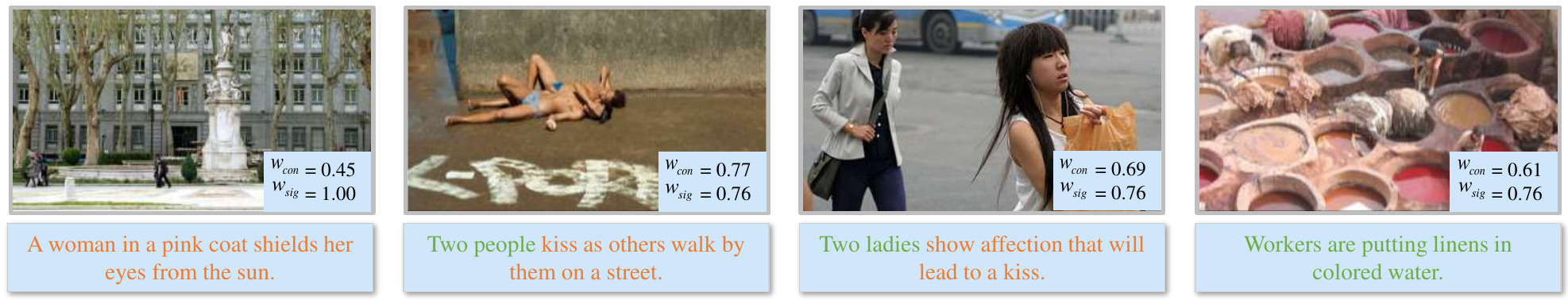} % Reduce the figure size so that it is slightly narrower than the column.
% \captionsetup{font=small}
\caption{Case study about vague samples on Flickr30K training set under 40\% noise. The correct and incorrect correspondences are marked in green and orange, respectively.
% \red{The vague samples have similar confidence weight $w_{con}$ and significance weight $w_{sig}$, but their correspondences differ greatly.}
}
\label{more_vague_sample}
\end{figure*}

% \begin{figure*}[t]
% \centering
% \includegraphics[width=1\textwidth]{figure/more_vague_sample.pdf} % Reduce the figure size so that it is slightly narrower than the column.
% % \captionsetup{font=small}
% \caption{Case study of more vague samples in Flickr30K training set under 40\% noise.}
% \label{more_vague_samples}
% \end{figure*}

\subsubsection{Reliability Analysis} The memory bank is used to assess the significance of samples, thus a reliable memory bank is a prerequisite for the superiority of our method. To verify the reliability of the memory bank, we visualize the data distribution of the clean dataset $D_c$ in Fig.~\ref{proportion_noisy_and_discard_clean}~(a). We compare different choices of threshold $\beta$ on the Flickr30K dataset with low (20\%) to high (60\%) noise ratios, respectively. From Fig.~\ref{proportion_noisy_and_discard_clean}~(a), it can be observed that the proportion of noisy samples in $D_c$ is larger when $\beta$ is small, but it can be negligible under a high threshold setting. Therefore, the samples in our memory bank are sufficiently reliable.
% and the significance weight is accurate.

In addition, we also visualize the distribution of samples discarded by SDD in Fig.~\ref{proportion_noisy_and_discard_clean}~(b) to verify the reliability of self-drop. Obviously, with the selection of an appropriate threshold, the self-drop strategy merely discards a small number of clean samples. At low noise ratios (20\%), the self-drop strategy discards only about 1\% of the clean samples, and at high noise ratios (60\%), the number of clean samples discarded by the self-drop strategy is even lower than 0.2\%. As we previously observed in Fig.~\ref{motivation}~(b), discarding a small portion of clean samples does not significantly impact the model, whereas a few noisy samples can have a substantial negative effect. Therefore, it is sufficiently reasonable that SDD filters noisy correspondence samples by self-drop.

\subsubsection{Similarity Distribution}
To further explore the influence of our method, we visualized the similarity distribution of clean and noisy pairs at different training stages of our SDD in Fig.~\ref{visualization_distribution} with the second-best baseline NPC for comparison.
As training progresses, the similarity of clean samples of SDD rises higher while that of noisy data decreases, exhibiting a clear trend of separation. This phenomenon indicates that our approach less fits noisy samples and remarkably suppresses the negative effect caused by them. Meanwhile, one could observe that there is a clear distinction between noisy and clean samples in NPC's distribution in the initial stage (Fig.~\ref{visualization_distribution}~(a)). However, the overlap between noisy and clean samples increases in the following epochs (Fig.~\ref{visualization_distribution}~(b)-(d)), indicating that NPC inevitably fits noisy data. Conversely, the noise samples and clean samples in SDD exhibit a more evident separation trend. This phenomenon accounts for the reason why SDD outperforms NPC across multiple datasets and demonstrates greater stability.

\subsubsection{Hyperparameter Analysis} The selection of hyperparameters $\alpha$ and $\beta$ is crucial for SDD. The former determines the threshold for discarding noisy samples, while the latter sets the threshold for constructing the memory bank. As illustrated in Fig.~\ref{visualization_distribution}~(a), the similarity of noisy samples is typically below 20, whereas that of clean samples is generally above 30. This phenomenon accounts for the hyperparameter setting $\alpha=20$ and $\beta=30$. In other words, $\alpha$ and $\beta$ are easy to adjust according to similarity distribution and will not complicate our approach.

\subsubsection{Case Study} We conducted the case study and show some examples intuitively. Fig.~\ref{case_study}~(a)-(d) respectively illustrate the cases of our qua-partition: noisy, vague, clean yet insignificant, and clean and significant. We report similarity score $sim$ for the noisy sample and the confidence weight $w_{con}$ and significance weight $w_{sig}$ for others.
In particular, noisy image-text pairs with low similarity are directly discarded, and thus their weights are not displayed. From Fig.~\ref{case_study}~(a), we can see that the content of images has no relationship with their corresponding captions. Therefore, it is a cost-effective choice to discard such samples directly. Vague samples are tough to distinguish from noisy data according to image-text similarity and they are assigned proper confidence weights (Fig.~\ref{case_study}~(b)). Clean samples share high $w_{con}$ while $w_{sig}$ balances their contribution to model training (Fig.~\ref{case_study}~(c)-(d)). 

Due to the complexity of vague samples, we present a case study about more vague samples in Fig.~\ref{more_vague_sample} for a comprehensive understanding of them. For the vague samples in Fig.~\ref{more_vague_sample}, we leverage different colors to distinguish between correct (green) and incorrect (orange) correspondences. Specifically, vague mismatched pairs may exhibit subtle semantic misalignment, \eg, action and the scene in the second and third images. Moreover, complete misalignment and proper alignment are also very common in vague  samples. Consequently, vague samples are located near the decision boundary~\cite{feng2023learning}, which presents a challenging issue to resolve in noisy correspondence. Therefore, we assign a smaller confidence weight to cautiously utilize these samples.

\section{Conclusion}
In this paper, we proposed a simple yet effective self-drop and dual-weight method to address a significant and challenging problem of learning with noisy correspondence. Specifically, we analyzed the effect of noisy and clean data pairs and found that for vision-language pre-training models, a small number of clean samples is more valuable than a majority of noisy ones. Based on this observation, we employed self-drop to discard potentially noisy samples to effectively mitigate the impact of noise. In addition, we adopted a dual-weight strategy to ensure that the model focuses more on significant samples while appropriately leveraging vague ones. Comprehensive experimental results revealed that our proposed approach surpasses contemporary state-of-the-art methods, yielding robust and competitive outcomes even under elevated noise ratios.

\bibliographystyle{IEEEtran}
\bibliography{arxiv}
% \bibitem{ref1}
% {\it{Mathematics Into Type}}. American Mathematical Society. [Online]. Available: https://www.ams.org/arc/styleguide/mit-2.pdf

% \bibitem{ref2}
% T. W. Chaundy, P. R. Barrett and C. Batey, {\it{The Printing of Mathematics}}. London, U.K., Oxford Univ. Press, 1954.

% \bibitem{ref3}
% F. Mittelbach and M. Goossens, {\it{The \LaTeX Companion}}, 2nd ed. Boston, MA, USA: Pearson, 2004.

% \bibitem{ref4}
% G. Gr\"atzer, {\it{More Math Into LaTeX}}, New York, NY, USA: Springer, 2007.

% \bibitem{ref5}M. Letourneau and J. W. Sharp, {\it{AMS-StyleGuide-online.pdf,}} American Mathematical Society, Providence, RI, USA, [Online]. Available: http://www.ams.org/arc/styleguide/index.html

% \bibitem{ref6}
% H. Sira-Ramirez, ``On the sliding mode control of nonlinear systems,'' \textit{Syst. Control Lett.}, vol. 19, pp. 303--312, 1992.

% \bibitem{ref7}
% A. Levant, ``Exact differentiation of signals with unbounded higher derivatives,''  in \textit{Proc. 45th IEEE Conf. Decis.
% Control}, San Diego, CA, USA, 2006, pp. 5585--5590. DOI: 10.1109/CDC.2006.377165.

% \bibitem{ref8}
% M. Fliess, C. Join, and H. Sira-Ramirez, ``Non-linear estimation is easy,'' \textit{Int. J. Model., Ident. Control}, vol. 4, no. 1, pp. 12--27, 2008.

% \bibitem{ref9}
% R. Ortega, A. Astolfi, G. Bastin, and H. Rodriguez, ``Stabilization of food-chain systems using a port-controlled Hamiltonian description,'' in \textit{Proc. Amer. Control Conf.}, Chicago, IL, USA,
% 2000, pp. 2245--2249.

% \end{thebibliography}

\end{document}